\documentclass[lettersize,journal]{IEEEtran} 
\usepackage{amsmath,amsfonts}
\usepackage[ruled,vlined]{algorithm2e}
\usepackage{array}
\usepackage[caption=false,font=normalsize,labelfont=sf,textfont=sf]{subfig}
\usepackage{textcomp}
\usepackage{stfloats}
\usepackage{url}
\usepackage{verbatim}
\usepackage{graphicx}
\usepackage{cite}
\usepackage{multirow}
\usepackage{flushend,cuted}
\begin{document}

\title{Binarizing by Classification: Is Soft Function Really Necessary?}

\author{Yefei He, Luoming Zhang, Weijia Wu, and Hong Zhou
\thanks{The authors are with the Key Laboratory for Biomedical Engineering of Ministry of Education, Zhejiang University, Hangzhou 310027, China (e-mail: billhe@zju.edu.cn; zluoming@zju.edu.cn; weijiawu@zju.edu.cn; zhouh@mail.bme.zju.edu.cn).}
\thanks{This work was supported by the National Key Research and Development Program of China (Grant No. 2022YFC3602601).}

}

\markboth{IEEE Transactions on Circuits and Systems for Video Technology,~Vol.~14, No.~8, NOVEMBER~2022}%
{Shell \MakeLowercase{\textit{et al.}}: A Sample Article Using IEEEtran.cls for IEEE Journals}


\maketitle

\begin{abstract}
Binary neural networks leverage $\mathrm{Sign}$ function to binarize weights and activations, which require gradient estimators to overcome its non-differentiability and will inevitably bring gradient errors during backpropagation. Although many hand-designed soft functions have been proposed as gradient estimators to better approximate gradients, their mechanism is not clear and there are still huge performance gaps between binary models and their full-precision counterparts. To address these issues and reduce gradient error, we propose to tackle network binarization as a binary classification problem and use a multi-layer perceptron (MLP) as the classifier in the forward pass and gradient estimator in the backward pass. Benefiting from the MLP's theoretical capability to fit any continuous function, it can be adaptively learned to binarize networks and backpropagate gradients without any prior knowledge of soft functions. From this perspective, we further empirically justify that even a simple linear function can outperform previous complex soft functions. Extensive experiments demonstrate that the proposed method yields surprising performance both in image classification and human pose estimation tasks. Specifically, we achieve 65.7\% top-1 accuracy of ResNet-34 on ImageNet dataset, with an absolute improvement of 2.6\%. Moreover, we take binarization as a lightweighting approach for pose estimation models and propose well-designed binary pose estimation networks SBPN and BHRNet. When evaluating on the challenging Microsoft COCO keypoint dataset, the proposed method enables binary networks to achieve a mAP of up to $60.6$ for the first time. Experiments conducted on real platforms demonstrate that BNN achieves a better balance between performance and computational complexity, especially when computational resources are extremely low.
\end{abstract}


\section{Introduction}
\IEEEPARstart{R}{ecently}, deep neural networks (DNNs) have achieved great success in many fields, including computer vision, natural language processing and multi-modality tasks. DNNs usually have massive parameters and high computational complexity. For example, ResNet-50\cite{he2016deep} has more than $25$M parameters and requires $4.1$G floating-point calculations for one inference. Deploying such a huge model requires powerful computing resources and mass storage. However, with the development of mobile phones and Internet of things (IoT) devices, there is an increasing demand for deploying neural networks on edge devices. This not only protects data privacy but also reduces the need for network transmission. On these devices with limited storage and computing power, excessive model parameters and computations are unacceptable. To solve this problem, various model compression and acceleration methods have been proposed, such as model quantization\cite{zhang2018lq,gong2019differentiable,liu2022ecoformer, zhuang2019structured, xu2022improving,nguyen2020layer}, pruning\cite{kang2019accelerator,zhao2021exploring,guo2020model, he2018soft, liu2018rethinking, molchanov2019importance}, distillation\cite{chen2021distilling,hinton2015distilling,guo2021distilling,jiao2019tinybert}, etc. Among them, binary neural network (BNN) compresses both weights and activations to 1-bit, which greatly reduces the model size and computational complexity. Meanwhile, the most frequently used Multiply Accumulate (MAC) operation during inference can be replaced by bit-wise operations like $\mathrm{XNOR}$ and $\mathrm{bitcount}$, which is faster and more power-saving than ordinary floating-point operations. As reported in XNOR-Net ~\cite{rastegari2016xnor}, BNN brings $32\times$ storage compression and up to $58\times$ computing acceleration, making it suitable for deployment on edge devices.

However, BNN greatly limits the model capacity and representational capability\cite{bulat2020high}, resulting in a huge degradation in network accuracy. For example, the pioneering work XNOR-Net ~\cite{rastegari2016xnor} only achieved  $51.2\%$ classification accuracy on ImageNet\cite{deng2009imagenet} dataset, with an accuracy gap of $18\%$ compared to the full-precision counterpart. To improve binary network performance, many follow-up studies have been proposed to minimize binarization error\cite{bulat2019xnor++,rastegari2016xnor,zhou2016dorefa}, design binary-friendly architecture\cite{liu2020bi, liu2020reactnet,zhu2019binary,liu2022rbnet} and better training algorithms\cite{IRNET,gong2019differentiable}. Among them, how to overcome the non-differentiability of sign function is the key to the optimization of binary networks. Generally, the $\mathrm{Sign}$ function is used to binarize the weight and activation, converting continuous floating-point values to discrete \{-1, +1\}. However, the exact gradient after this operation is zero almost everywhere. To address this, straight-through estimator (STE)~\cite{bengio2013estimating} is widely used to train quantized networks, which approximate gradients with $\mathrm{Identity}$ function. Obviously, this will bring huge gradient error and the error will accumulate with the backpropagation process, causing great damage to the model accuracy. In order to reduce the gradient error, DSQ ~\cite{gong2019differentiable} first proposes to approximate the gradient with a derivable soft function, thus helping the optimization of quantized networks. Many follow-up works ~\cite{IRNET,ding2022ie,zhang2022root} also propose improved soft functions to further improve the accuracy of BNNs. However, the working mechanism of soft function is still not clear. For instance, is it better if soft functions are closer to $\mathrm{Sign}$ function in shape? If so, there is little difference from directly using the $\mathrm{Sign}$ function. Moreover, the increasingly complicated form of soft functions limits their further development, which makes us wonder: Does soft function have to be so complex to perform well?

\begin{strip}  \tiny \centering
Copyright © 2023 IEEE. Personal use of this material is permitted. However, permission to use this material for any other purposes must be obtained from the IEEE by sending an email to pubs-permissions@ieee.org.
\end{strip}\raggedend 

\begin{figure*}[thb]
    \centering
    \includegraphics[width=\columnwidth]{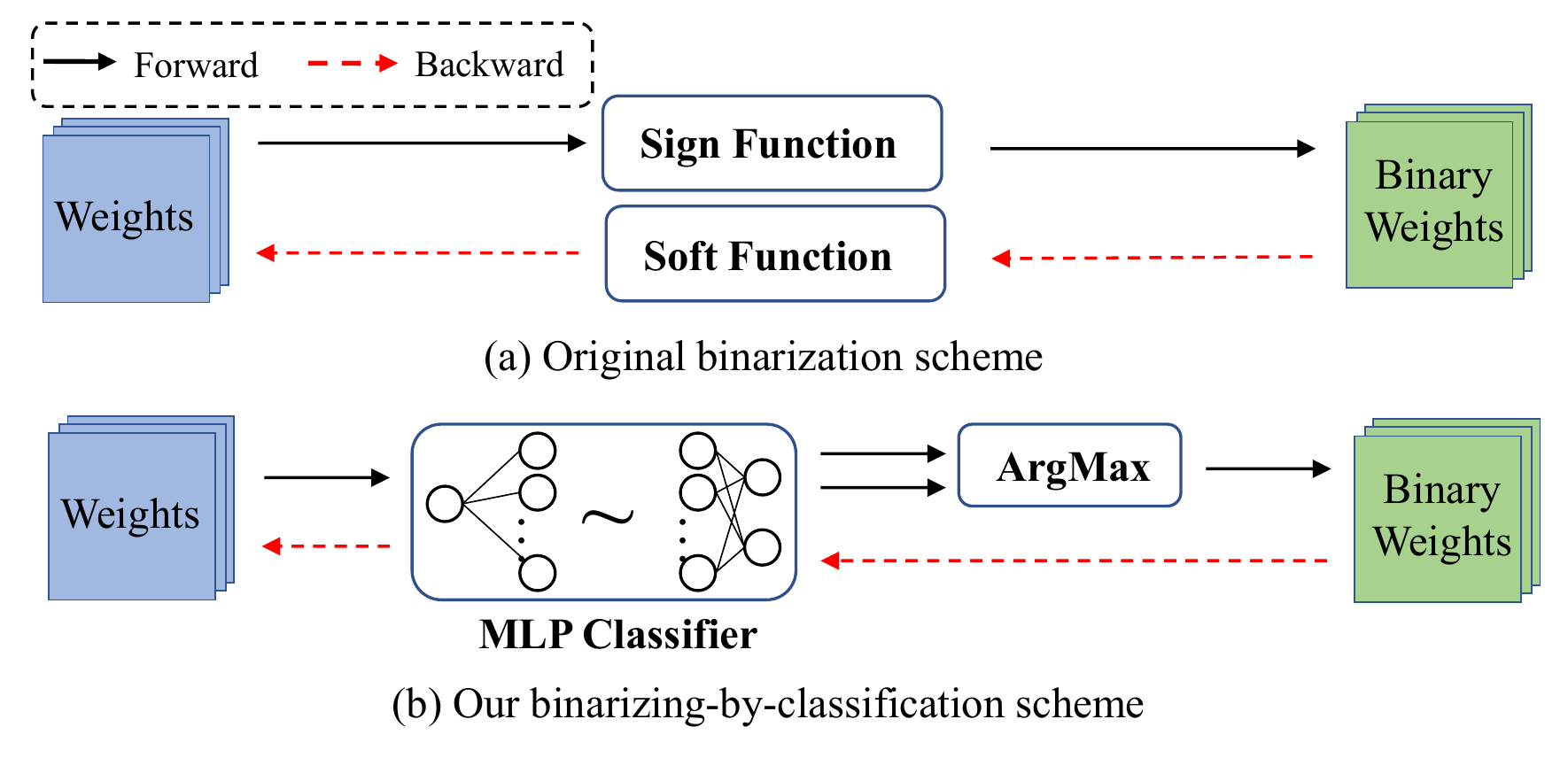}
    \caption{\textbf{The weight binarization and gradient approximation processes of original binarization scheme and our Binarizing-by-Classification scheme.} "$ \sim $" in the MLP classifier represents the optional activation layer. The same MLP is used in both forward and backward process.}
    \label{fig:bbc}
\end{figure*}

To solve the above question, this paper proposes to treat model binarization as a binary classification problem and use an MLP as the binary classifier to binarize weights in the forward pass and approximate gradients in the backward pass, as illustrated in Figure~\ref{fig:bbc}. 
Using MLP instead of a specific soft function has the following advantages.
Firstly, the same MLP is applied in both forward and backward pass, greatly reducing the forward-backward mismatch. As mentioned above, traditional methods utilize $\mathrm{Sign}$ function to binarize in the forward pass and soft functions to approximate gradient in the backward pass, which inevitably brings mismatch and gradient error. By contrast, we use the same MLP in the forward and backward passes, which is inherently derivable and can well propagate gradients; 
Secondly, MLP is proved to be able to approximate any continuous function~\cite{hornik1991approximation}, so that we do not have to specify a functional forms for gradient estimator. In previous studies, functional forms are hand-crafted and only few variables are mutable during the training process. Instead, the whole MLP we propose is learnable in the proposed method, which could approximate any continuous functions including previously proposed soft functions. Therefore, the model capacity of the gradient estimator is greatly enlarged. Moreover, the width, depth and structure of MLP can be modified so that we can verify the optimal complexity required by the gradient estimator.

After training, we use the MLP to binarize and fix the weights in the model, so that the MLP can be discarded and does not generate additional computations during the inference. 
With the proposed powerful Binarizing-By-Classification (BBC) module, this paper further expands the application of BNN to more complex vision tasks besides image classification, such as human pose estimation on large-scale Microsoft COCO 2017\cite{coco} dataset. To our best knowledge, we are the pioneering work to test BNNs performance over human pose estimationin tasks in real applications.
 
We summarize our contributions as follows:
\begin{itemize}
\item We propose a novel scheme to treat binarization as a binary classification problem and use an MLP as the binary classifier. The MLP-based classifier is adaptively learned to binarize weights of the network, which greatly improves its performance.
\item Benefiting from the strong generalization performance of MLPs, we compare and analyze MLPs of different complexity with complex soft functions proposed in previous work. Experiment results show that complex soft functions are not necessary for optimization, even a simple linear function is able to outperform previous state-of-the-art (SOTA) methods.
\item For the first time, we evaluate the performance of BNN on large-scale Microsoft COCO keypoint dataset. Combined with the proposed BBC module, BNNs outperform current SOTA lightweighting methods. Benefiting  from the small size and the fast inference speed of the BNNs, it provides a new approach for the lightweighting of pose estimation models.
\end{itemize}

\section{Related Work} \label{sec:relatedwork}
\begin{figure}[thb]
    \centering
    \includegraphics[width=\columnwidth]{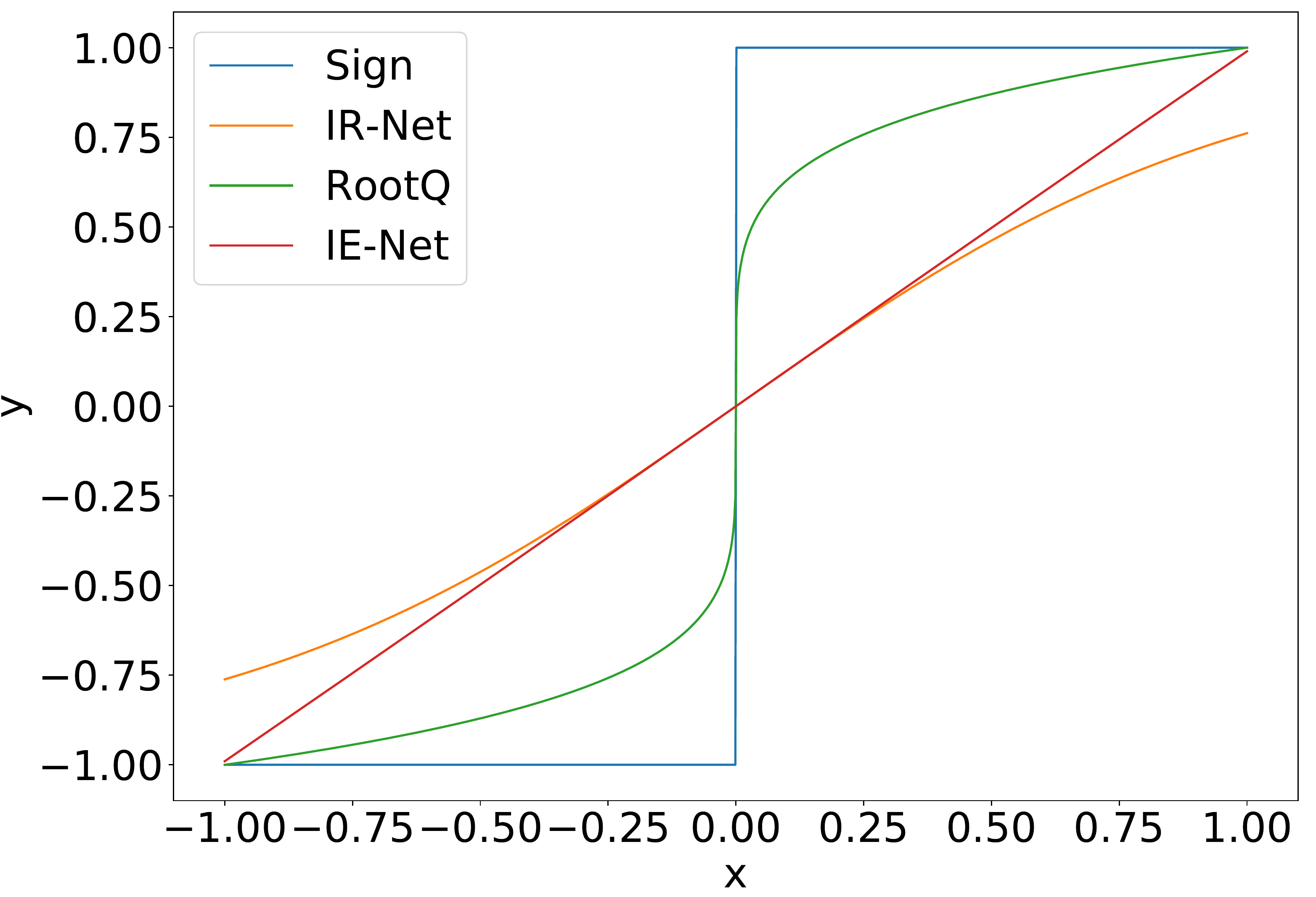}
    \caption{\textbf{Comparisons of different gradient approximation functions.} IR-Net: $y=k\mathrm{Tanh}(tx)$;  RootQ: $y=\frac{\mathrm{Sign}(x)}{2}(2x)^\alpha$; IE-Net: $y=r(-\mathrm{Sign}(x)\frac{3q^2x^2}{4}+\sqrt{3}qx)$}
    \label{fig:functioncomparison}
\end{figure}

Network binarization quantizes both weights and activations to 1-bit, thus greatly reducing the memory footprint and computational cost. It is first proposed in BNN~\cite{hubara2016binarized}, which proves the practicability of training BNNs on small image classification datasets like MNIST\cite{deng2012mnist} and CIFAR-10\cite{Krizhevskycifar10}. However, the performance degraded severely when larger datasets are encountered, such as ImageNet~\cite{deng2009imagenet}. To reduce the accuracy gap between BNNs and their full-precision counterparts, many follow-up studies have been proposed, ranging from binary-friendly architecture designs~\cite{liu2020reactnet,liu2020bi,zhuang2019structured, liu2022rbnet}, optimizer selection~\cite{liu2021adam, alizadeh2019systematic} and new training schemes~\cite{real2b, zhuang2018towards}. Among them, the optimization of BNN is the key problem. To overcome the non-differentiability of $\mathrm{Sign}$ function, STE\cite{bengio2013estimating} and many soft functions ~\cite{IRNET, gong2019differentiable, ding2022ie,le2022adaste} have been proposed to estimate the gradient of $\mathrm{Sign}$ function. Also, a better gradient estimation is orthogonal to other binarization methods and can be applied together. However, the forms of these proposed functions are complex, as shown in Figure~\ref{fig:functioncomparison}. It’s difficult to explain the advantages of one function over another. To alleviate this problem, MetaQuant~\cite{chen2019metaquant} explicitly uses a meta-network to learn the gradient of the network, splitting the forward and backward process. On the contrary, we binarize the network and calculate gradients implicitly with one MLP, which serves as binary classifier in the forward pass and gradient estimator in the backward pass. LNS~\cite{han2020training} calculates weight groups together with different meta-networks and introduces extra supervision to train the meta-network, while we only use one MLP shared through the whole network without extra supervision. Neural Encoder~\cite{l2b} uses a neural encoder to binarize the network. However, it is not well explained and not compared to previous soft functions. In this paper, we want to explore whether such complex soft functions are really necessary with the help of MLP modules. Using binary-friendly network structures\cite{liu2020bi,liu2020reactnet} and two-stage training schemes\cite{real2b,zhuang2018towards} are proven to be very helpful, which have brought great accuracy improvement to BNNs. In this paper, we also prove that our optimization module can co-exist with these methods, and the combination of the two can further improve the accuracy of binary networks.

Although BNNs have made great progress in image classification tasks, there are few studies that applied BNN to more challenging visual tasks, such as object detection\cite{wang2020bidet,xu2021layer}, human pose estimation\cite{bulat2019improved} and semantic segmentation\cite{zhuang2019structured}. Compared with image classification tasks, it is even harder to achieve good performance due to the high information requirements of these tasks and the limited representation ability of BNNs. For instance, a recent study~\cite{bulat2019improved} proposed an improved binary network for pose estimation. However, the method is only evaluated on small datasets like MPII\cite{andriluka20142d} and there is a huge gap between this method and the full precision model. In this paper, we further evaluated our method on the large-scale Microsoft COCO Keypoint dataset~\cite{coco} and test it on real platforms, proving that BNN can achieve competitive performance in complex visual tasks and provide a new approach for the lightweighting of models on these tasks.

The literature closely related to our work includes~\cite{efficientCNN4pose, liu2022rbnet, xu2022improving}. The work of ~\cite{efficientCNN4pose} proposes to rely on lightweight CNN architectures to achieve a good accuracy-efficiency trade-off on human pose estimation tasks, which is a common approach. In contrast, our work pioneers the use of BNN as a lightweight approach for pose estimation networks, which can provide better accuracy with lower complexity. While STTN~\cite{xu2022improving} enables the model to automatically determine the threshold for quantization, it is only suitable for ternary networks. On the other hand, BNNs are more difficult to optimize but can achieve higher compression rates and computing efficiency. RB-Net~\cite{liu2022rbnet} proposes improving the accuracy of BNNs through a new binary-friendly structure and evaluated it only on image classification tasks. In contrast, our goal is to reduce the gradient error brought by the $\mathrm{Sign}$ function, which is the key problem of BNN optimization. Additionally, we extend the application of BNNs to pose estimation tasks and test them on real platforms. To the best of our knowledge, few studies have tested the performance of BNNs under complex tasks on real platforms.

\section{Preliminaries}
BNNs represent both weights and activations with only one bit. To accomplish this, the $\mathrm{Sign}$ function is frequently utilized to binarize real values:
\begin{equation}
\hat{x}=\mathrm{Sign}(x)=\left\{
\begin{aligned}
+1&, if x\ge0, \\
-1&, otherwise.
\end{aligned}
\right.
\end{equation}

However, the non-differentiability of the $\mathrm{Sign}$ function is problematic during backward pass. As a solution, the Straight-Through Estimator (STE)~ is commonly used to address this issue:
\begin{gather}
    \frac{\partial \mathcal{L}}{\partial x}\approx\left\{
    \begin{aligned}
    \frac{\partial \mathcal{L}}{\partial \hat{x}}&, \mathrm{if}\ \lvert x \rvert \le1 \\
    \noindent 0&, \mathrm{otherwise}.
    \end{aligned}
    \right.
\end{gather}

To further approximate real values, scaling factors $\alpha$ are introduced to reduce binarization error.
\begin{equation}
\label{eq:scale}
\alpha = \frac{{\lVert \mathbf{x} \rVert}_{\ell_1}}{n},
\end{equation}

Since all parameters and activations in BNN are represented by a single bit, the most frequently used Multiply Accumulate (MAC) operations can be replaced by bit-wise operations ($\mathrm{XNOR}$ and $\mathrm{bitcount}$), which can be formulated as:
\begin{equation}
o=\mathrm{bitcount}(\mathrm{XNOR}(\mathbf{w}, \mathbf{a}))
\end{equation}
where o is the result of MAC operation, $\mathbf{w}$ indicates a weight vector and $\mathbf{a}$ indicates an activation vector. In practice, both vectors are stored in 64-bit registers. Compared with floating-point MAC operations, BNNs delivery an inference acceleration by $64\times$ and memory savings by $32\times$.

\section{Methodology}
\subsection{Binarizing By Classification} \label{BBC}
Currently, BNNs are challenging to quantize in a post-training manner, and they can only be trained or fine-tuned to restore their accuracy. To address the problem that the $\mathrm{Sign}$ function is not differentiable, STE~\cite{bengio2013estimating} and various types of gradient estimators~\cite{IRNET, ding2022ie, zhang2022root} are utilized to propagate the gradient through binarization. Despite these gradient estimators being based on different function forms, they share a common motivation that these functions gradually approximate the $\mathrm{Sign}$ function as the training progresses. Therefore, they are similar in shape and control the degree of approximation of the $\mathrm{Sign}$ function by adjusting their parameters. Figure~\ref{fig:functioncomparison} provides a visual comparison between these functions. Although these methods improve the accuracy of BNNs, the mechanism and motivation behind the functional forms are still unclear. For instance, it is challenging to explain why the gradient estimator based on the $\mathrm{Root}$ function outperforms its $\mathrm{Tanh}$-based counterpart. Moreover, the increasing complexity of function forms limits the development of more accurate gradient estimators.

To break free from this paradigm, we approach the problem from a new perspective. Researches~\cite{hornik1991approximation, hornik1989FFN} have shown that Feed-Forward Networks can approximate any continuous function and square integrable function. Therefore, instead of designing a specific function form as in previous research, we propose a novel Binarizing-by-Classification (BBC) scheme. Specifically, we incorporate an MLP $M_{\theta}$, which is parameterized by $\theta$ and shared in the whole model, to serve as binarizer in the forward pass and gradient estimator in the backward pass. The output dimension of MLP is set to $2$, making it a binary classifier for weight binarization.

Formally, in the forward pass, we take MLP as binarizer to binarize weights instead of $\mathrm{Sign}$ function: 
\begin{equation}
    \Tilde{W}= M_{\theta}(W). 
\end{equation}

As $\mathrm{Argmax}$ is used to discretize values, elements in $\Tilde{W}$ will all be $0$ or $+1$. To maintain consistency with previous BNNs, a linear transformation $\mathrm{f}$ is then conducted to convert values from $\{0,1\}$ to $\{-1,1\}$: 
\begin{equation}
    \hat{W}=\mathrm{f}(\Tilde{W}) = 2\Tilde{W} - 1. 
\end{equation}

In the backward pass, the gradient $\textit{w.r.t}$ $\Tilde{W}$ can be easily derived because $\mathrm{f}$ is a linear transformation:
\begin{equation} \label{eq:gradwavew}
\frac{\partial \mathcal L}{\partial \Tilde{W}} =\frac{\partial \mathcal L}{\partial \hat{W}} \frac{\partial f(\Tilde{W})}{\partial \Tilde{W}}=2\frac{\partial \mathcal L}{\partial \hat{W}}.
\end{equation}

Since MLP is naturally differentiable, the gradient $\textit{w.r.t}$ $W$ then can be computed with our MLP-based gradient estimator $M_\theta$:
\begin{equation} \label{eq:gradw}
\frac{\partial \mathcal L}{\partial W} =\frac{\partial \mathcal L}{\partial \Tilde{W}} \frac{\partial \Tilde{W}}{\partial W}=\frac{\partial \mathcal L}{\partial \Tilde{W}} \frac{\partial M_\theta(W)}{\partial W}.
\end{equation}

To enable the MLP to estimate the optimal gradient, its parameter $\theta$ must evolve during training. Therefore, we set it as learnable and train it alongside the entire network. The gradient $\textit{w.r.t}$  $\theta$ can be derived by:

\begin{equation} \label{eq:gradtheta}
\frac{\partial \mathcal L}{\partial \theta} =\frac{\partial \mathcal L}{\partial \Tilde{W}} \frac{\partial \Tilde{W}}{\partial \theta}.
\end{equation}

Compared to prior studies, our method does not explicitly define a particular function form for the gradient estimator, thereby significantly increasing its capacity (\textit{i.e.}, both the function form and parameters are learnable). For instance, in the previous study RootQ~\cite{zhang2022root} (or IE-NET~\cite{ding2022ie}), only the variables $\alpha$ (or $r$ and $q$) evolve during training (refer to Figure~\ref{fig:functioncomparison}). The gradient estimator is constrained to utilizing the $\mathrm{Root}$ (or $\mathrm{Polynomial}$) function. In contrast, our approach employs a neural network as the gradient estimator, which is a more flexible and versatile method.

To construct an MLP for the gradient estimator, we have designed two structures with different complexity. The first structure consists of a simple input layer $M_{in}$ and an output layer $M_{out}$, without any nonlinear layer. In this case, the MLP is equivalent to a linear transformation:

\begin{equation}
    \Tilde{W} = M_{out}(M_{in}(W))
\end{equation}

The second structure builds upon the first by adding a nonlinear layer, $\sigma$, which can be any nonlinear activation layer. The MLP then can be expressed as:
\begin{equation}
    \Tilde{W} = M_{out}(\sigma(M_{in}(W)))
\end{equation}

In this study, we employ the $\mathrm{Tanh}$ and $\mathrm{ReLU}$ activation functions as nonlinearities $\sigma$. It is important to note that, even if the MLP uses $\mathrm{Tanh}$ as the activation function, it is not equivalent to directly using $\mathrm{Tanh}$ as the gradient estimator, as in the case of IR-Net~\cite{IRNET}. It has been widely studied that an MLP with sufficient capacity and non-linearity can approximate $\textbf{any continuous function}$ with high precision~\cite{hornik1991approximation, hornik1989FFN}. As we train the MLP in an end-to-end manner, it adaptively seeks the optimal gradient estimation. Therefore, the final state of the MLP depends on the specific training process and is not limited to a $\mathrm{Tanh}$ function.

Besides the non-linearity, the width and depth of the MLP also play a crucial role in the function fitting ability. This can be determined by the dimension of the hidden layer and the number of MLP layers. In this study, we also conduct ablation experiments to verify the impact of the complexity of the gradient estimator.

The training process of our BBC module is summarized in Algorithm~\ref{algor1}.

\begin{algorithm}[htbp]
	\caption{The training process of the network with proposed binarizing-by-classification.} 
	\label{algor1}
	\small
		\textbf{Require}: a binary classifier $\mathbf M_{\theta}$, a linear shift function $\mathrm{f}$, full-precision weights $\mathbf W \in\mathbb R^{n}$, the input data $\mathbf A \in\mathbb R^{n}$.\\
		{\textbf{Forward propagation}}\\
		\quad Compute binary weights$\hat{\mathbf W}$ by binary classifier:\\
		\quad \ForEach{$w_i$ in $\mathbf W$}{
    		\quad \quad $\Tilde{w_i} = \mathbf M_{\theta}(w_i)$\\
    		\quad \quad $\hat{w_i} = f(\Tilde{w_i})$\\}
		\quad $\hat{\mathbf W} = \{\hat{w_0}, \hat{w_1},..., \hat{w_n}\}$\\
		\quad Compute binary activations $\hat{\mathbf A}$:\\
		\quad \quad $\hat{\mathbf A} = Sign(\mathbf A)$\\
		\quad  Calculate the output:\\
		\quad \quad $\mathbf Z = \hat{\mathbf W} \odot \hat{\mathbf A} $\\
		
		{\textbf{Backward propagation}}\\
		\quad Compute the gradient of $\mathbf W$ according to Equation~\ref{eq:gradw}\\
		\quad Compute the gradient of $\theta$ according to Equation~\ref{eq:gradtheta}\\
		
		{\textbf{Update parameters}}\\
		\quad Update $\mathbf W$: $\mathbf W=\mathbf W-\eta_{w}\frac{\partial{\mathcal{L}}}{\partial{\mathbf W}}$\\
		\quad Update $ \theta $: $\theta=\theta-\eta_{\theta}\frac{\partial{\mathcal{L}}}{\partial{\theta}}$
\end{algorithm}

\subsection{Binary Networks For Human Pose Estimation}

\begin{figure*}[thb]
    \centering
    \includegraphics[width=\columnwidth]{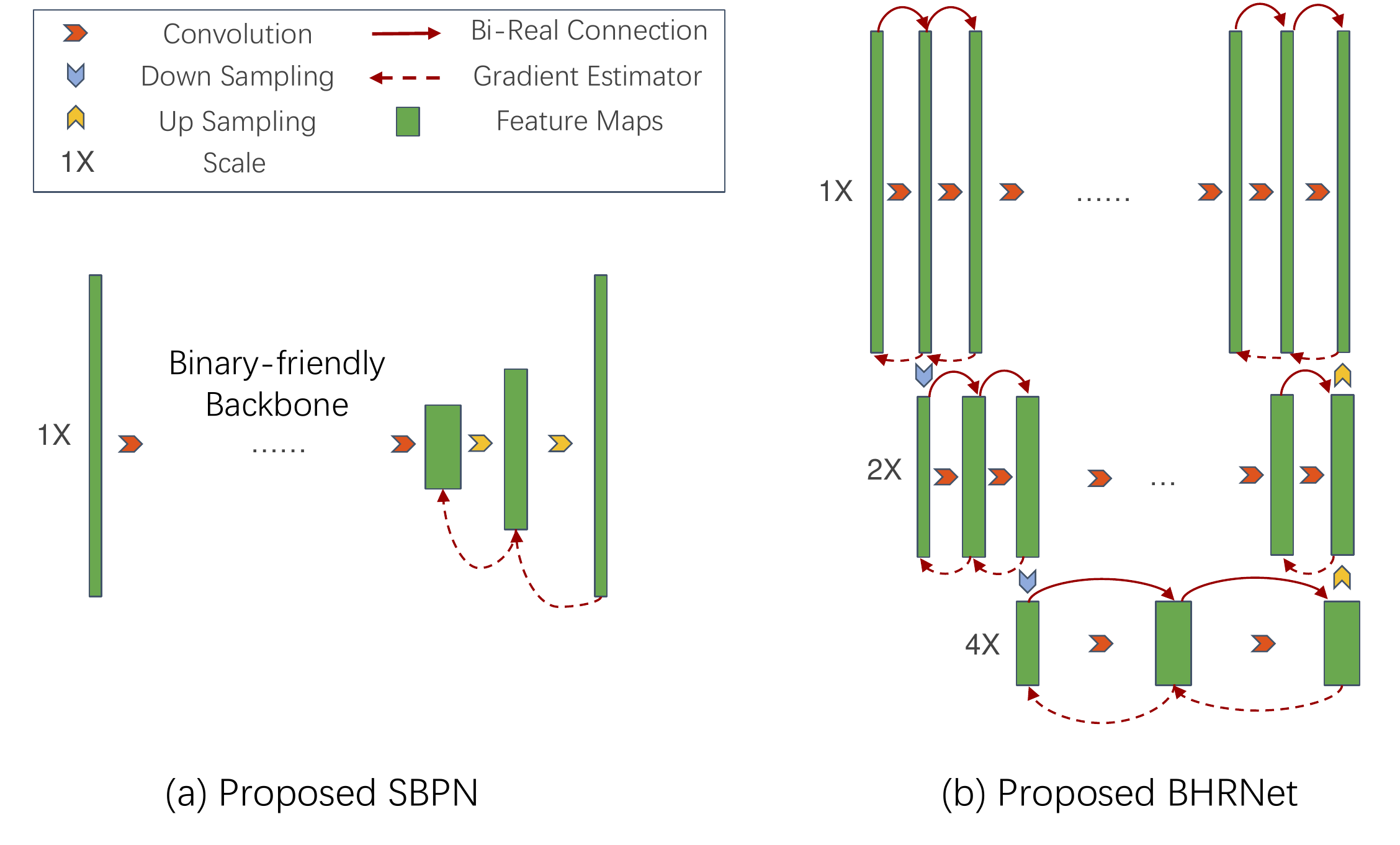}
    \caption{\textbf{The structure of the proposed SBPN and BHRNet.} To build a binary-friendly network for pose estimation, Bi-Real connection and BBC module are integrated with the backbone model. BHRNet further utilizes a pyramid structure to improve the performance.}
    \label{fig:bhrnet}
\end{figure*}

Although there have been many lightweight networks developed for human pose estimation~\cite{2021zhaoestimating, wang2022lite, neff2020efficienthrnet, efficientCNN4pose}, few attempts have been made to use BNN as the backbone network. In this section, we explore the feasibility of using network binarization as a lightweight approach.

First, we propose Simple Binary Pose Networks (SBPN), which are based on traditional binary-friendly network structures and directly connect several deconvolutional layers after the backbone for upsampling to obtain the final output, following the Simple-Baseline~\cite{simplebaseline}. Specifically, we adopt two binary-friendly network structures that have proven to be very effective, namely Bi-Real~\cite{liu2020bi} and ReActNet~\cite{liu2020reactnet}, as our backbones. Then we add deconvolutional layers on the feature map as the prediction head to get the final output, as shown in Figure~\ref{fig:bhrnet} (a). Combined with the proposed BBC module, the binary network backbone has strong feature extraction capabilities and can be optimized well during backward pass. Experimental results in later sections demonstrate that the proposed SBPN already outperform previous lightweight pose estimation methods, proving the ability of BNNs to maintain accuracy while greatly compressing the network. Although more than $80$\% of the parameters of the network have been binarized, the extra deconvolutional layers still uses floating-point calculations, which reduces the overall compression ratio. This is because the commonly used pyramid structure is not used in the method proposed by the Simple-Baseline~\cite{simplebaseline}, and only the last three deconvolutional layers are used for upsampling.

To overcome the aforementioned limitations and further enhance the accuracy of the model, it is crucial to maintain the pyramid structure while ensuring the network structure's binary-friendliness. To achieve this, we introduce the Binary High-Resolution Net (BHRNet). As demonstrated in Figure~\ref{fig:bhrnet} (b), we utilize the HigherHRNet~\cite{higherhrnet} as our backbone, which parallels high-to-low resolution subnetworks to efficiently extract features of different scales. To facilitate binarization, the convolution module used for feature extraction is transformed into a Bi-Real structure and employs the BBC module proposed earlier in both the forward and backward passes. The Bi-Real structure leverages the layer-wise shortcut to effectively preserve the information in the network, while the BBC module substantially reduces the gradient error during backpropagation caused by the $\mathrm{Sign}$ function. Compared with the SBPN, BHRNet has fewer parameters and a higher binarization ratio. By incorporating the above three enhancements, our proposed BHRNet achieves a breakthrough performance in BNN-based pose estimation, which will be elaborated in Section~\ref{sec:cocoresult}.


\section{Image classification experiment}
\subsection{Implementation Details}
\textbf{Datasets and network structure}: In this study, the image classification experiments are conducted on two standard benchmarks: CIFAR10 ~\cite{Krizhevskycifar10} and ImageNet (ILSVRC12) ~\cite{deng2009imagenet}. To prove the generality of our method, we employ it on widely-used network structures, including ResNet-20 ~\cite{he2016deep} for CIFAR10 and ResNet-18, ResNet-34 for ImageNet. All convolutional layers in networks are quantized to 1-bit except the first one, as is the common practice of BNN. The proposed MLP for approximating gradients is implemented with a one-layer linear model without non-linear parts, if not otherwise specified. We binarize both weights and activations because it is more challenging and sensitive to the binarization method, and bit-wise operations can be used only when both weights and activations are binarized. In Section~\ref{sec:comparesoft} where we compare our method with other soft functions, ResNet is implemented with Bi-Real structure as previous research~\cite{liu2020bi,IRNET} did for a fair comparison. Then in Section~\ref{sec:compareSOTA}, we integrate our BBC module with advanced BNN structures and training strategies to prove that the combination of them can further improve the accuracy.

\textbf{Training strategy}: For the experiments on CIFAR10 dataset, we apply SGD as our optimizer with a momentum of $0.9$ and a weight decay of $1e-4$. The initial learning rate is set to $0.1$ and the cosine annealing schedule is adopted to adjust the learning rate. The batchsize is set to $128$ and we train networks from scratch for $400$ epochs. For the experiments on ImageNet dataset, we use an Adam\cite{kingma2014adam} optimizer without weight decay. The initial learning rate is set to $2.5e-3$ and decreases linearly while training. Instead of using a 2-step training scheme\cite{real2b} (binarizing activations first then weights), the main results are achieved by training networks from scratch for $120$ epochs without intermediate training steps or warmups. In this way, we can compare with other soft functions more fairly. However, we also do a few experiments with the 2-step training scheme\cite{real2b} and improved binary structure\cite{liu2020reactnet} to prove that our method is able to achieve SOTA performance. In this case, we utilize their official implementations and follow their settings for a fair comparison. The CIFAR10 and ImageNet experiments are conducted with 1 RTX3090 GPU and 4 RTX3090 GPUs, respectively.

\subsection{Comparison with Soft Functions} \label{sec:comparesoft}
\subsubsection{CIFAR10}
We first evaluate our method on CIFAR10 dataset by comparing it with existing SOTA binary quantization methods, including DSQ ~\cite{gong2019differentiable}, DoReFa ~\cite{zhou2016dorefa}, IR-Net ~\cite{IRNET}, NE ~\cite{l2b} and SD-BNN ~\cite{xue2022self}. Among them, DSQ ~\cite{gong2019differentiable} and IR-Net ~\cite{IRNET} propose to use $\mathrm{Tanh}$ as soft functions to approximate the $\mathrm{Sign}$ function. SD-BNN also adopts the same soft function as IR-Net. As shown in Table\ref{cifarresult}, our BBC outperforms previous methods with complex soft functions and further narrows the performance gap between the binary and full-precision model (less than $3.8\%$ over ResNet-20). It should be noted that we only use a MLP without any non-linearity as the gradient estimator, which indicates that a simple linear function can be competent to approximate the gradient, even though $\mathrm{Tanh}$ is closer to the $\mathrm{Sign}$ function in shape. We will further verify this view in Section~\ref{sec:ablation}. 

\begin{table}[h]
    \caption{Accuracy comparison with soft functions on CIFAR10 dataset.}
    \label{cifarresult}
\centering
\begin{tabular}{llll}
\hline
Model                      & Method        & Bit-Width (W/A) & Acc.(\%)      \\ \hline
\multirow{8}{*}{ResNet-20} & FP            & 32/32          & 91.7          \\ 
                           & DoReFa~\cite{zhou2016dorefa}        & 1/1            & 79.3          \\ 
                           & DSQ~\cite{gong2019differentiable}           & 1/1            & 84.1          \\ 
                           & NE~\cite{l2b}           & 1/1            & 85.3          \\ 
                           & LNS~\cite{han2020training}           & 1/1            & 85.8          \\ 
                           & IR-Net~\cite{IRNET}        & 1/1            & 86.5          \\ 
                           & SD-BNN~\cite{xue2022self}       & 1/1            & 86.9          \\ 
                           & \textbf{Ours} & \textbf{1/1}   & \textbf{87.9} \\ \hline
\end{tabular}
\end{table}

\subsubsection{ImageNet}
To further verify the robustness of the method, we evaluate it on the large-scale ImageNet\cite{deng2009imagenet} dataset. Table~\ref{imagenetresult} shows the comparison with current SOTA binary methods over ResNet-18 and ResNet-34. It can be seen that our BBC achieves the best accuracy in both network structures. Our result over ResNet-18 is even comparable to IR-Net over ResNet-34 ($62.3\%$ vs. $62.9\%$), with only half of the parameters. The absolute accuracy increase brought by BBC can be up to $2.6\%$ over ResNet-34 ($65.7\%$ vs. $63.1\%$). The experimental results prove that our BBC consistently outperforms the existing methods that utilize hand-crafted soft functions.

\begin{table}[h]
    \caption{Accuracy comparison with soft functions on ImageNet dataset.}
    \label{imagenetresult}
\centering
\begin{tabular}{llll}
\hline
Model                      & Method        & Bit-Width (W/A) & Acc.(\%)      \\ \hline
\multirow{7}{*}{ResNet-18} & FP            & 32/32          & 69.6          \\ 
                           & XNOR-Net~\cite{rastegari2016xnor}      & 1/1            & 51.2          \\  
                           & Bi-Real~\cite{liu2020bi}      & 1/1            & 56.4          \\  
                           & IR-Net~\cite{IRNET}        & 1/1            & 58.1          \\  
                           & LNS~\cite{han2020training}        & 1/1            & 59.4          \\  
                           & NE ~\cite{l2b}           & 1/1            & 61.3          \\ 
                           & \textbf{Ours} & \textbf{1/1}  & \textbf{62.3} \\ 
 \hline
\multirow{5}{*}{ResNet-34} & FP            & 32/32          & 73.3          \\  
                           & Bi-Real~\cite{liu2020bi}       & 1/1            & 62.2          \\  
                           & IR-Net~\cite{IRNET}         & 1/1            & 62.9          \\ 
                           & NE~\cite{l2b}           & 1/1            & 63.1   
                           \\
                           & \textbf{Our}  & \textbf{1/1}   & \textbf{65.7} \\ \hline
\end{tabular}
\end{table}


\subsection{Combination with SOTA BNNs} \label{sec:compareSOTA}
As we analyzed in Section~\ref{sec:relatedwork}, a good gradient estimator is orthogonal to other BNN optimization methods, such as architecture design and training strategies, and can be combined together to further improve the accuracy. To demonstrate the coexistence ability of our method with prior work, we combine BBC with activation binarization method\cite{tu2022adabin}, 2-stage training scheme\cite{real2b} and binary-friendly network structure~\cite{liu2020reactnet} respectively. Among them, we use the ResNet-18 model structure for the comparison with \cite{tu2022adabin} and \cite{real2b}. The results are reported in Table~\ref{SOTAresult}. After replacing original soft functions or STE\cite{bengio2013estimating} in these methods, the combination of the BBC module and these recent BNN optimization methods consistently narrow the gap between the full-precision and binary model, proving the flexibility and generality of our method.

\begin{table}[h]
    \caption{Accuracy comparisons for other BNN methods on ImageNet dataset.}
    \label{SOTAresult}
\centering
\begin{tabular}{cccc}
\hline
Method Type                                                                               & Method      & BitWidth (W/A) & Acc. (\%) \\ \hline
\multirow{3}{*}{\begin{tabular}[c]{@{}c@{}}Activation\\ Binarization\end{tabular}} 
& FP      & 32/32            &  69.6       \\
& AdaBin\cite{tu2022adabin}      & 1/1            &  63.1         \\
                                                                                   & \textbf{+ Ours}       & \textbf{1/1}            &  \textbf{63.4}         \\ \hline
\multirow{3}{*}{\begin{tabular}[c]{@{}c@{}}2-stage\\ Training\end{tabular}}        
& FP      & 32/32            &  69.6       \\
& Real-to-Bin\cite{real2b} & 1/1            &  65.4         \\
                                                                                   & \textbf{+ Ours}       & \textbf{1/1}            &  \textbf{66.1}         \\ \hline
\multirow{3}{*}{\begin{tabular}[c]{@{}c@{}}Advanced\\ Structure\end{tabular}}      
& FP      & 32/32            &  72.4       \\
& ReActNet-A\cite{liu2020reactnet}    & 1/1            &     69.4      \\
                                                                                   & \textbf{+ Ours}       & \textbf{1/1}            &    \textbf{69.9}        \\ \hline
\end{tabular}
\end{table}

\subsection{Ablation Analysis} \label{sec:ablation}
In this subsection, we investigate the relationship between the complexity of soft function and its performance. Previous work has proposed a lot of complex soft functions and tried to analyze their mechanism. Among various functions, $\mathrm{Tanh}$ function is widely used for its similar shape to $\mathrm{Sign}$ function. DSQ and IR-Net held the view that the soft function needs to evolve during training, and it coincides with the $\mathrm{Sign}$ function eventually. But is this really the case? We validate this point of view through the adaptive MLP-based gradient estimator. As mentioned above, MLP with sufficient capacity can approximate any continuous function. Therefore, we change the nonlinearity of MLP by adding activation layers and increase its complexity by increasing the number of MLP layers. The activation layer is located after the first linear layer if there is one.

\begin{table}[h] 
\caption{The performance of MLP-based classifier under different settings.}
\label{tab:ablation}
\resizebox{\columnwidth }{!}{
\begin{tabular}{lllll}
\hline
Model                      & Number of Layers & Number of Channels & Activation Layer & Acc.(\%)       \\ \hline
\multirow{5}{*}{ResNet-20} & 0                & -                  & -             & 83.32          \\ 
                           & \textbf{1}       & \textbf{-}         & \textbf{-}       & \textbf{87.98} \\ 
                           & 1                & -                  & Tanh             & 87.87          \\ 
                           & 2                & 100                & -                & 87.70          \\ 
                           & 2                & 100                & Tanh             & 40.05          \\ \hline
\end{tabular} }
\end{table}

The experimental results are shown in Table~\ref{tab:ablation}. Unexpectedly, a single-layer neural network without any activation layer achieves the best accuracy. In this case, the MLP is actually equivalent to a linear function, and the gradient is simply multiplied by the coefficients of the linear function when backpropagating. As the complexity and non-linearity of the MLP increase, the accuracy of the model decreases slightly. When the MLP has both hidden layers and $\mathrm{Tanh}$ activation layers, the model fails to converge. We assume this is due to overfitting caused by the high complexity of the MLP without strong supervision.

\begin{figure}[ht]
    \centering
    \includegraphics[width=\columnwidth]{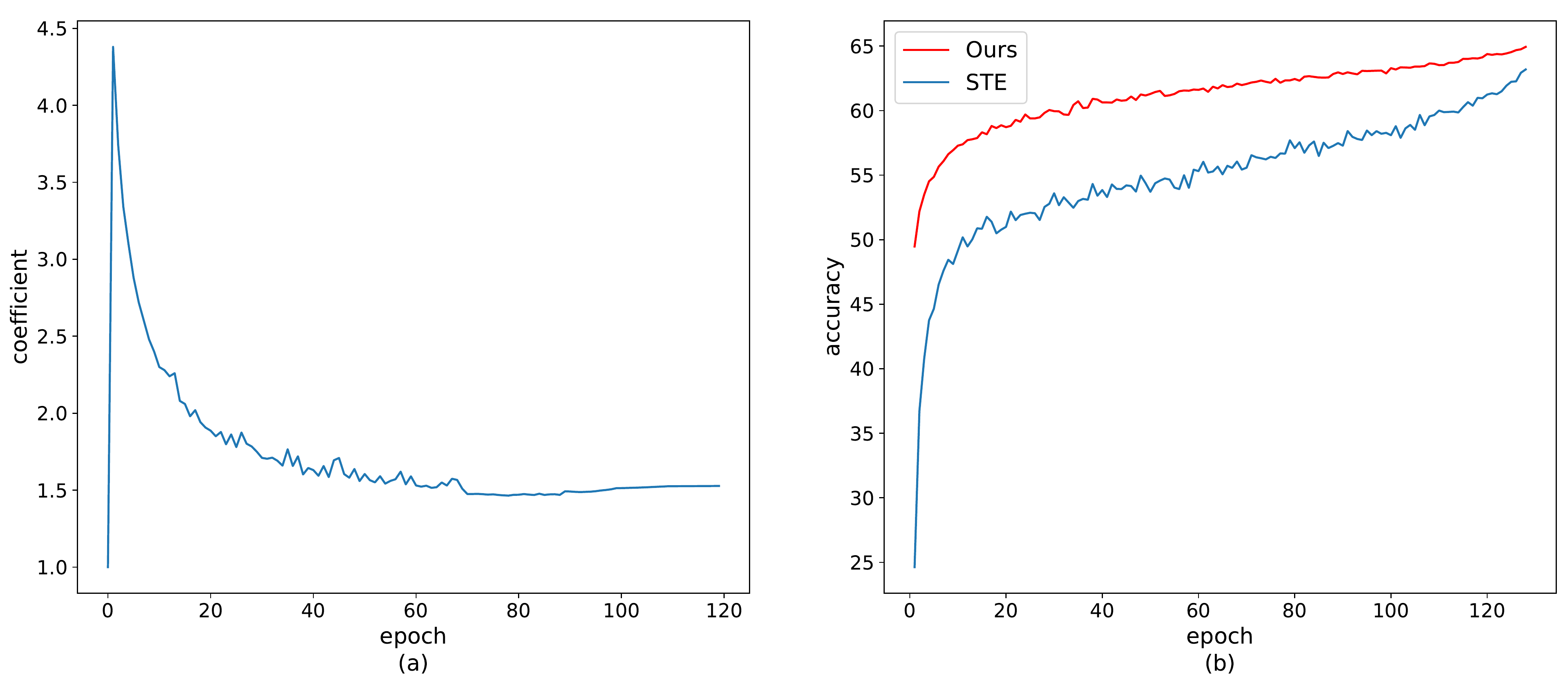}
    \caption{\textbf{Evolution process of the coefficient during training (a) and how this evolution helps convergence (b).}}
    \label{fig:evolution}
\end{figure}

\subsection{Evolution of MLP-based Gradient Estimator}
Since the single-layer MLP without any activation layer (\emph{i.e.}, a linear function) achieves the best accuracy, we conduct experiments to see how the coefficient of this linear function changes. It should be noted that this coefficient is equal to the sum of two neurons of the single-layer MLP. Figure~\ref{fig:evolution} (a) plot the coefficient per epoch during training.

We can see that that coefficient increases rapidly at the beginning of training, which means it's rapidly approaching the $\mathrm{Sign}$ function. After that, the coefficient slowly decreased and stabilized around $1.5$. The adaptive coefficient also helps the convergence of BNNs, as shown in Figure~\ref{fig:evolution} (b). Overall, training BNNs with our BBC module has faster convergence, higher accuracy and a more stable training process. Given the fact that STE is equivalent to a linear soft function where the coefficient is exactly $1$, the MLP behaves similarly to STE when it's stabilized. This shows that in many cases STE only needs to multiply an adaptive coefficient to work well, and there is no need to design a complex soft function to replace it. The idea is similar to AdaSTE\cite{le2022adaste}, which also propose a data-dependent STE method. However, it follows the Mirror Descent\cite{beck2003mirror} approach and manually designs functions and rules for calculating key parameters, while our method evolves automatically during training and achieves better performance.

\subsection{Computational Complexity Analysis}
Compared with vanilla BNNs, an MLP-based gradient estimator is proposed in this paper to help binarization. Therefore, additional parameters and computational overhead will be introduced. However, this additional cost is almost negligible and even smaller than the previous method.

In the training phase, the additional computation depends on the complexity of MLP. Since we have demonstrated in the section~ that a single-layer neural network achieves the best accuracy, we use this configuration in the rest of the experiments. In this case, the extra parameters and calculations from MLP are almost negligible, as shown in Table~\ref{tab:bbccomplexity}. The gradient estimator (with one linear layer and no activation layer) in our method only includes two neurons, where the additional calculations are even fewer than soft functions like DSQ\cite{gong2019differentiable} and IR-Net\cite{IRNET}. For each weight, we only need to multiply twice and compare values, while soft functions need to calculate nonlinear functions (like $\mathrm{Tanh}$) and multiple multiplications (for coefficients). 

\begin{table}[h] 
\caption{Complexity comparison of different gradient estimators. We count the additional calculation required for each weight element.}
\label{tab:bbccomplexity}
\resizebox{\columnwidth }{!}{
\begin{tabular}{ccc}
\hline
Method & Number of multiplications & Number of non-linear operations \\ \hline
DSQ~\cite{gong2019differentiable}    & 2                        & 1                               \\
IR-Net~\cite{IRNET} & 2                        & 1                               \\
IE-Net~\cite{ding2022ie} & 4                        & 2                               \\
RootQ~\cite{zhang2022root}  & 3                        & 2                               \\
\textbf{Ours}   & \textbf{2}               & \textbf{0}                      \\ \hline
\end{tabular} }
\end{table}

In the inference phase, there is no additional computation, parameters or memory consumption at all. For activations, we maintain the same binarization method as prior work that simply uses $\mathrm{Sign}$ function. For binary weights, they can be fixed after the network training, so the MLP can be totally discarded during inference, keeping the same computational complexity as previous methods. 

\section{Human pose estimation experiment}
\subsection{Experiment Settings}
We evaluate our method on the large-scale Microsoft COCO dataset~\cite{coco}, which contains more than $110000$ images labeled with $17$ keypoints for human pose estimation. Models are trained only on the $train$ set and we report the performance on the $val$ set. The input resolution is $384\times288$ and $512\times512$ for SBPN and BHRNet, respectively. For data augmentation and hyper-parameter settings, we follow the previous work~\cite{simplebaseline, higherhrnet}. It’s worth noticing that no pretrained models are used for all experiments, and we train all models from scratch with the one-step training scheme. For SBPN and BHRNet, the BBC module proposed in section~\ref{BBC} is adopted to correct gradients. 

\subsection{Main Results} \label{sec:cocoresult}
Binarizing networks is an alternative way to compress the pose estimation model, which can greatly reduce the total OPs of the model. We compare our method with SOTA lightweight human pose estimation models. As shown in Table~\ref{cocoresult}, directly combining binary-friendly network structures with deconvolution layers will bring a lot of floating-point operations (refer to $4_{th}$ and $6_{th}$ rows of Table~\ref{cocoresult}), reducing the compression ratio brought by binarization. On the contrary, our proposed BHRNet maintains the multi-scale detection ability for pose estimation, while making the model structure binary friendly, outperforming the current SOTA method with fewer OPs (only $7.93$G). As far as we know, it's the first time that a binary network achieves a result of $60.6$ mAP on the challenging Microsoft COCO dataset.

\begin{table}[h]
    \caption{Results on Microsoft COCO $val$ set. Here, "FLOPs" means floating point operations, "BOPs" means binary operations and "OPs" means total operations, which is calculated by OPs = BOPs / 64 + FLOPs.}
    \label{cocoresult}
\centering
\begin{tabular}{lllll}
\hline
Model                            & FLOPs         & BOPs           & OPs            & AP            \\ \hline
HigherHRNet ~\cite{cheng2020higherhrnet}                      & 47.9G         & -              & 47.9G          & 69.9          \\
Lightweight OpenPose ~\cite{osokin2018real}             & 18.0G         & -              & 18.0G          & 42.8          \\
EfficientHRNet-H\_{-2} ~\cite{neff2020efficienthrnet}                   & 15.8G         & -              & 15.8G          & 52.9          \\
SBPN (Bi-Real based) & 10.30G        & 3.69G          & 10.36G         & 54.1          \\
LitePose-S ~\cite{wang2022lite}                       & 10.0G         & -              & 10.0G          & 56.8          \\
SBPN (ReAct based)   & 11.17G        & 10.37G         & 11.33G         & 58.7          \\
\textbf{BHRNet}                  & \textbf{7.3G} & \textbf{40.2G} & \textbf{7.93G} & \textbf{60.6} \\ \hline
\end{tabular}
\end{table}

\subsection{Deployment Efficiency}
In this section, we deploy binarized and other lightweight pose estimation networks on an actual platform to test their performance in real-world applications. All experiments are conducted on an RK3588 SoC, which is equipped with a quad-core 2.4GHz ARM Cortex-A76 and a quad-core 1.8GHz ARM Cortex-A55 CPU. To fully utilize the acceleration provided by binarization, we use the high-performance inference framework Bolt~\cite{bolt2022}, which optimizes both binary and floating-point computation. To ensure a fair comparison, we use single-threaded computation for all models.

During deployment, BNNs are packed, stored and computed in 1-bit format, significantly reducing the model size and memory usage. With the efficient XNOR-bitcount bitwise operations, the inference speed can also be significantly improved compared to full-precision networks with similar complexity. As shown in Table VI, the proposed SBPN (ReAct-based) achieves higher accuracy than the SOTA method LitePose-S, while enjoying a $1.7\times$ actual computation acceleration and only requiring $50\%$ memory usage, achieving an ideal balance between accuracy and computational complexity. Compared to LitePose-XS, the proposed SBPN (Bi-Real-based) achieves significant performance advantages ($54.1$ vs. $40.6$) with only slightly increased computational complexity and memory consumption, proving the performance advantages of BNN under extremely low resources. Moreover, the proposed BHRNet can achieve an AP of more than $60$ on the COCO2017 dataset, achieving the SOTA performance of the lightweight pose estimation network with only a model size of $9.2$M. The real deployment of BNNs demonstrates that binary pose estimation networks can be extended to complex visual tasks and achieve a perfect balance between performance and computational complexity. In particular, BNN can significantly improve the performance compared with traditional lightweight method in the platform with extremely low resource. 

Moreover, it is important to note that general computation platforms like CPU and GPU have all bit, integer and floating-point computation unit. Although binary networks utilize bit operations to achieve faster processing, it may not be optimized on traditional computing platforms such as CPU or GPU. However, despite this limitation, we still demonstrate superior compression and acceleration capabilities of BNN compared to traditional lithgweight methods on general ARM CPU platforms. On FPGA or ASIC platforms, it can greatly simplify circuit design and achieve better performance and energy consumption ratios since most calculations can be done with bit operations~\cite{BCNNFPGA, liang2018fpbnn}.

\begin{table}[]
    \caption{Comparison of resourse consumption and time cost of different light-weighted pose estimation network. Here, ``Model Size"" refers to the size of storage space occupied by model files and ``Memory Usage"" denotes the amount of memory used during model inference.}
    \label{cocodeploy}
    \centering
\resizebox{\columnwidth }{!}{
\begin{tabular}{ccccc}
\hline
Model                                     & \begin{tabular}[c]{@{}c@{}}Model Size\\ (MB)\end{tabular} & \begin{tabular}[c]{@{}c@{}}Memory Usage\\ (MB)\end{tabular} & \begin{tabular}[c]{@{}c@{}}Latency\\ (ms)\end{tabular} & AP   \\ \hline
LitePose-XS                               & 6.4                                                       & 33.8                                                        & 97.1                                                   & 40.6 \\
EfficientHRNet-H\_{-2}                  & 32.7                                                      & 139.0                                                       & 809.1                                                  & 52.9 \\
LitePose-S                                & 10.5                                                      & 94.9                                                        & 390.1                                                  & 56.8 \\ \hline
\textbf{SBPN (Bi-Real based)} & 9.7                                                       & 40.5                                                        & 154.8                                                  & 54.1 \\
\textbf{SBPN (ReAct based)}   & 15.6                                                      & 46.7                                                        & 226.0                                                  & 58.7 \\
\textbf{BHRNet}                           & 9.2                                                       & 78.7                                                        & 499.4                                                  & 60.6 \\ \hline
\end{tabular} }
\end{table}

\section{Conclusion \& Discussion} \label{sec:conclusion}
In this paper, we propose a solution to address the non-differentiability of the $\mathrm{Sign}$ function when training accurate BNNs. Specifically, we propose a BBC scheme that binarizes networks with an MLP-based binary classifier in the forward pass, which then acts as a gradient estimator during the backward pass. Leveraging the powerful generalization ability of MLP, we demonstrate that designing complex soft functions as gradient estimators is suboptimal for training BNNs. Our experiments show significant accuracy improvements on ImageNet by using a simple MLP-based gradient estimator, which is equivalent to a linear function. Moreover, we propose two binary networks, SBPN and BHRNet, for the challenging task of human pose estimation. These approaches combine binary-friendly network structures with the BBC gradient estimation module and achieve SOTA performance on the COCO keypoint dataset. Compared with previous lightweight networks for pose estimation, our proposed BHRNet achieves better performance with less computational costs, demonstrating the effectiveness of BNNs in complex vision tasks.

In the future, we will extend BNN with BBC module to other downstream vision tasks such as segmentation and dense detection. Also, in this paper we limit the application of quantizing-by-classification mechanism to binary neural networks (\emph{i.e.}, binary classification). This mechanism is theoretically feasible with multi-bit quantization as well. This will be investigated in future work.

\begin{IEEEbiography}[{\includegraphics[width=1in,height=1.25in,clip,keepaspectratio]{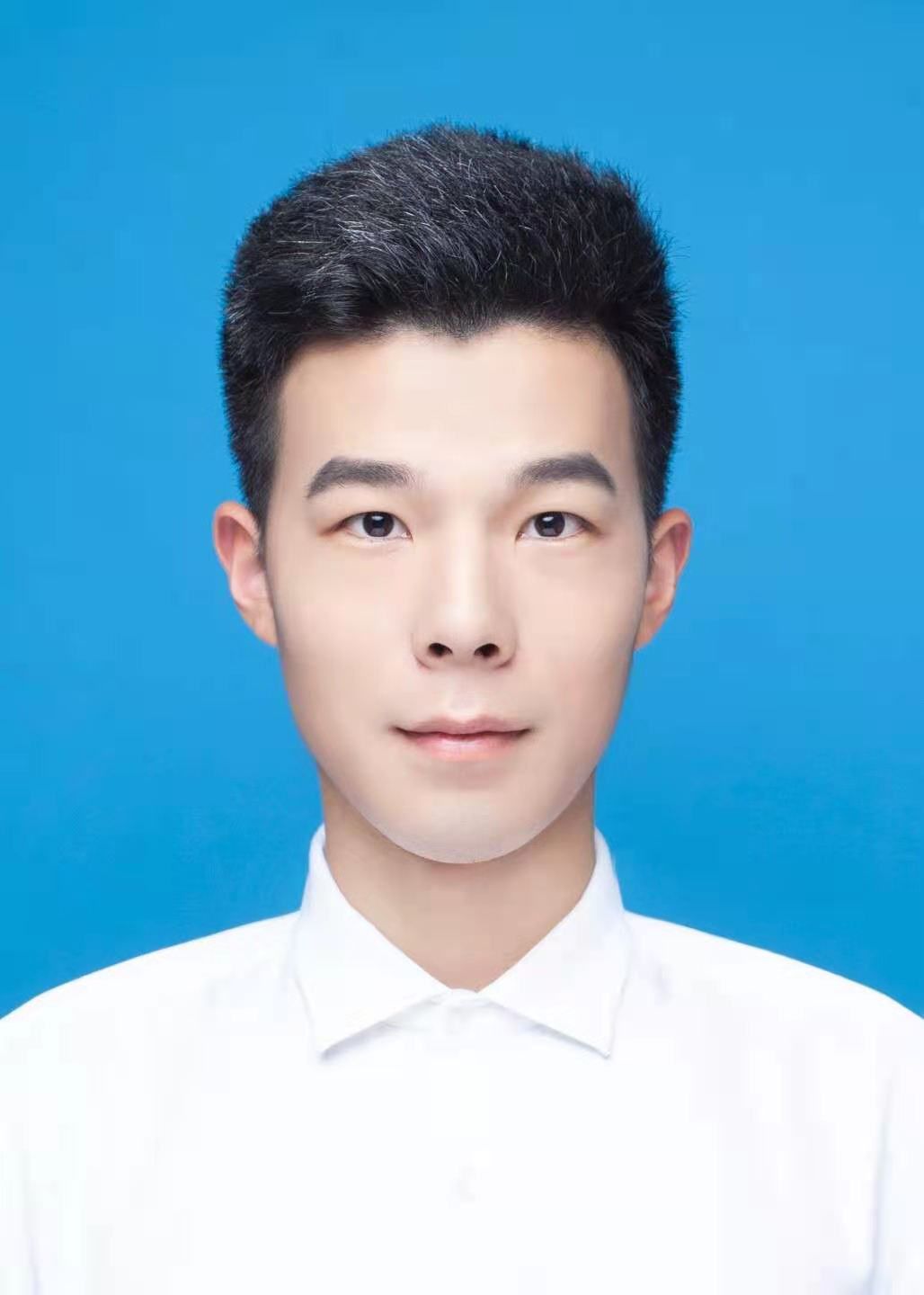}}]{Yefei He} received the B.E. degree in instrument science from Zhejiang University, Hangzhou, China, in 2021. He is currently pursuing the Ph.D. degree with the College of Biomedical Engineering \& Instrument Science, Zhejiang University, Hangzhou, China. His research interests revolve around computer vision, network compression, and acceleration techniques.
\end{IEEEbiography}
\begin{IEEEbiography}[{\includegraphics[width=1in,height=1.25in,clip,keepaspectratio]{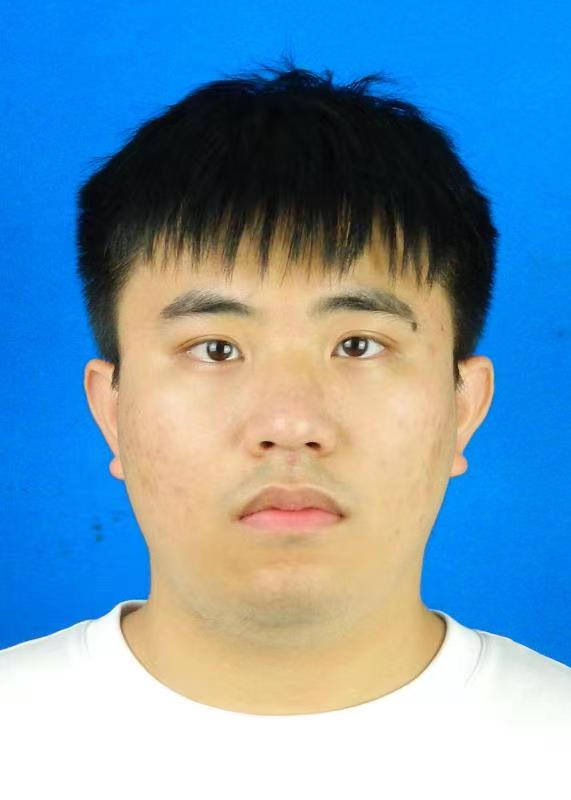}}]{Luoming Zhang} received the B.E. degree in measurement and control technology and Instrumentation Program Control from Zhejiang University, Hangzhou, China, in 2019. He is currently working toward Ph.D. degree with the College of Biomedical Engineering \& Instrument Science, Zhejiang University, Hangzhou. His research interests include computer vision and model compression.
\end{IEEEbiography}
\begin{IEEEbiography}[{\includegraphics[width=1in,height=1.25in,clip,keepaspectratio]{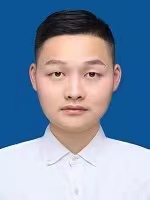}}]{Weijia Wu} is a fourth-year PhD student at Zhejiang University and a visiting student at the National University of Singapore. His research interests are quite relevant to long-form video understanding, scene text detection, recognition and cross-modal video-and-language retrieval.
\end{IEEEbiography}
\begin{IEEEbiography}[{\includegraphics[width=1in,height=1.25in,clip,keepaspectratio]{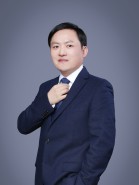}}]{Hong Zhou} received the B.E. degree in measurement technology and instrument from Zhejiang University, Hangzhou, China, in 1995, and the Ph.D. degree in instrument science and technology from Zhejiang University, Hangzhou, China, in 2000. He is currently a Professor with the College of Biomedical Engineering \& Instrument Science, Zhejiang University. His current major research interests include video analysis technology and embedded system.
\end{IEEEbiography}



\vfill

\end{document}